\documentclass{article}
\usepackage{spconf,amsmath,graphicx,epsfig}
\usepackage{verbatim}
\usepackage{xcolor}
\usepackage{amsfonts,todonotes}
\usepackage[algoruled,boxed,lined]{algorithm2e}
\usepackage{balance}
\usepackage{fancyhdr}
\usepackage{hyperref}
\makeatletter
\def\BState{\State\hskip-\ALG@thistlm}

\title{Compact Graph Architecture for Speech Emotion Recognition}
\name{Amir Shirian, Tanaya Guha}
\address{University of Warwick, UK}
\begin{document}
\ninept
\maketitle
\thispagestyle{fancy}
\renewcommand{\headrulewidth}{0pt}
\lhead{The paper will appear in ICASSP 2021}

\begin{abstract}
We propose a deep graph approach to address the task of speech emotion recognition. A compact, efficient and scalable way to represent data is in the form of graphs. Following the theory of graph signal processing, we propose to model speech signal as a cycle graph or a line graph. Such graph structure enables us to construct a Graph Convolution Network (GCN)-based architecture that can perform an accurate graph convolution in contrast to the approximate convolution used in standard GCNs. We evaluated the performance of our model for speech emotion recognition on the popular IEMOCAP and MSP-IMPROV databases. Our model outperforms standard GCN and other relevant deep graph architectures indicating the effectiveness of our approach. When compared with existing speech emotion recognition methods, our model achieves comparable performance to the state-of-the-art with significantly fewer learnable parameters ($\sim$30K) indicating its applicability in resource-constrained devices. Our code is available at \href{https://github.com/AmirSh15/Compact\_SER}{/github.com/AmirSh15/Compact\_SER}.
\end{abstract}
\section{Introduction}
\label{sec:intro}
Machine recognition of emotional content in speech is crucial in many human-centric systems, such as behavioral health monitoring 
and empathetic conversational systems. 
Speech Emotion Recognition (SER) in general is a challenging task due to the huge variability in emotion expression and perception across speakers, languages and cultures.

Many SER approaches follow a two-stage framework, where a set of Low-Level Descriptors (LLDs) are first extracted from raw audio. The LLDs are then input to a deep learning model to generate discrete (or continuous) emotion labels \cite{tang2018end,zhao2018exploring,huang2016attention,mirsamadi2017automatic}. While using hand-crafted acoustic features is still common in SER, lexical features  \cite{aldeneh2017pooling, jin2015speech} and
log Mel spectrograms are also used as inputs \cite{mao2019deep}. Spectrograms are often used with Convolutional Neural Networks (CNNs) \cite{mao2019deep} that does not explicitly model the speech dynamics. Explicit modeling of the temporal dynamics is important in SER as it reflects the changes in emotion dynamics \cite{han2018towards}. To capture the temporal dynamics of emotion, recurrent models, especially the Long-Short Term Memory networks (LSTMs), are popular \cite{zhao2018exploring,huang2016attention,mirsamadi2017automatic}. The recurrent models, though predominant in SER, often lead to complex architecture with millions of trainable parameters.

A compact, efficient and scalable way to represent data is in the form of graphs. Graph Convolutional Networks (GCNs) \cite{kipf2017semi} have been successfully used to address various problems in computer vision and natural language processing, such as action recognition \cite{yan2018spatial}, object tracking \cite{gao2019graph} and text classification \cite{yao2019graph}. In the context of audio analysis, we are aware of only one recent work that proposed an attention-based graph neural network architecture for few-shot audio classification \cite{zhang2019few}. 

Motivated by the recent success of GCNs, we propose to adopt a deep graph approach to SER. We base our work on \textit{spectral} GCNs which have a strong foundation on graph signal processing \cite{6494675}. Spectral GCNs perform convolution operation on the spectrum of the graph Laplacian considering the convolution kernel (a diagonal matrix) to be learnable \cite{bruna2013spectral}. This involves eigen decomposition of the graph Laplacian matrix, which is computationally expensive. To reduce the computational cost, ChebNet approximates the convolution operation (including the learnable convolution kernel) in terms of Chebyshev polynomials \cite{defferrard2016convolutional}. The most popular form of GCN uses a first order approximation of the Chebyshev polynomial to further simplify the convolution operation to a linear projection \cite{kipf2017semi}. Such GCN models are simple to implement, and have been successfully used for various node classification tasks in social media networks and citation networks \cite{kipf2017semi}.

In this paper, we cast SER as a graph classification problem. We model a speech signal as a simple graph, where each node corresponds to a short windowed segment of the signal. Each node is connected to only two adjacent nodes thus transforming the signal to a \emph{line graph} or a \emph{cycle graph}. Owing to this particular graph structure, we take advantage of certain results in graph signal processing \cite{ortega2018graph} to perform accurate graph convolution (in contrast to the approximations used in popular GCNs). This leads to a light-weight GCN architecture with superior emotion recognition performance on the IEMOCAP \cite{busso2008iemocap} and the MSP-IMPROV \cite{GideonKADP17} databases. 

To summarize, the contributions of our work are as follows: (i) To the best of our knowledge, this is the first work that takes a graph classification approach to SER. (ii) Leveraging theories from graph signal processing, we propose a GCN-based graph classification approach that can efficiently perform accurate graph convolution. (iii) Our model has significantly fewer trainable parameters ($\sim$30K only). Despite its smaller size, our model achieves superior performance on both IEMOCAP and MSP-IMPROV databases outperforming relevant and competitive baselines.

\section{Proposed Approach}
\label{sec:proposed}
\begin{figure*}[tb]
\centering
   \fbox{\includegraphics[width=1.0\linewidth, trim=0mm 7mm 0mm 4mm, clip=true]{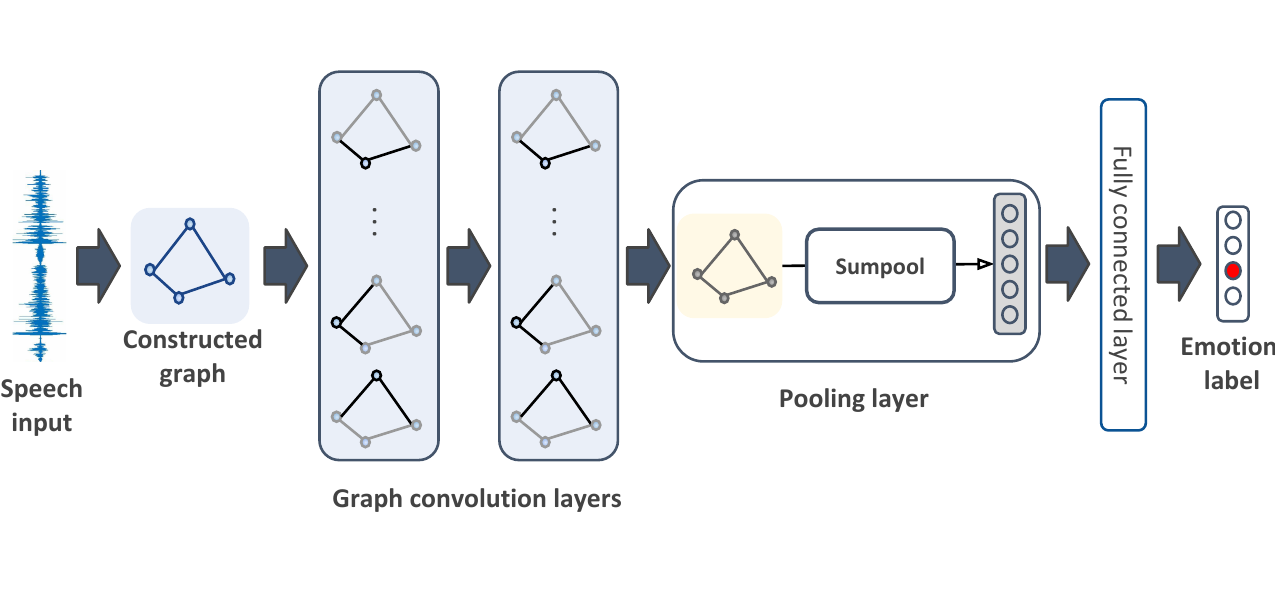}}
  \caption{Our proposed graph-based architecture for SER consists of two graph convolution layers and a pooling layer to learn graph embedding from node embeddings to facilitate emotion classification.}
  \label{fig:graph_con}
\end{figure*}
In this section, we describe our graph classification approach to SER. First, we construct a graph from each speech sample. Next, we develop a new GCN architecture that assigns a discrete emotion label to each (speech sample transformed to) graph. Fig.~\ref{fig:graph_con} gives an overview of our approach. Below, we describe each component in detail.
\subsection{Graph construction}
Given a speech signal (utterance), the first step is to construct a corresponding graph $\mathcal{G} = (\mathcal{V},\mathcal{E})$, where $\mathcal{V}\in\{v_i\}_{i=1}^M$ is the set of $M$ nodes, and $\mathcal{E}$ is the set of all edges between the nodes. The adjacency matrix of $\mathcal{G}$ is denoted by $\mathbf{A}\in\mathbb{R}^{M\times M}$ where an element $(\mathbf{A})_{ij}$ denotes the edge weight connecting $v_i$ and $v_j$.

\begin{figure}[t]
    \begin{minipage}[t]{0.9\linewidth}
    \centering
    \includegraphics[width=7.3cm,height=2.8cm, trim = 0mm 3mm 0cm 0cm, clip=true]{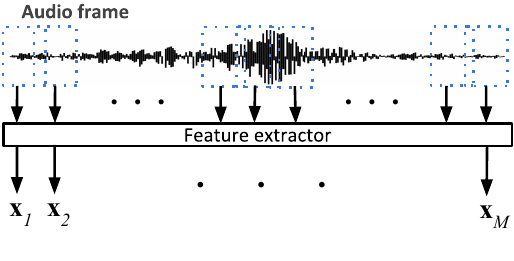}
    (a)
  \end{minipage}
  
  \begin{minipage}[t]{0.45\linewidth}
    \centering
    \includegraphics[width=1.0\linewidth, trim = 0mm 0cm 0cm 0cm, clip=true]{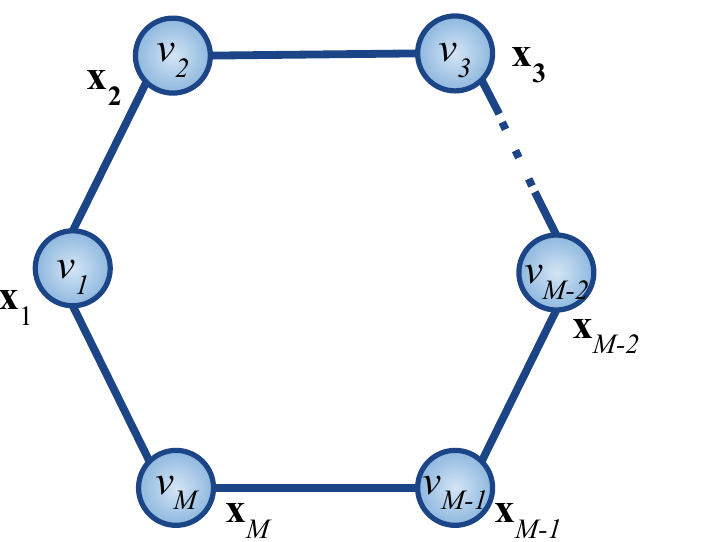}
    (b)
  \end{minipage}
  \begin{minipage}[t]{0.45\linewidth}
    \centering
    \includegraphics[width=1.0\linewidth, trim = 0mm 0cm 0cm 0cm, clip=true]{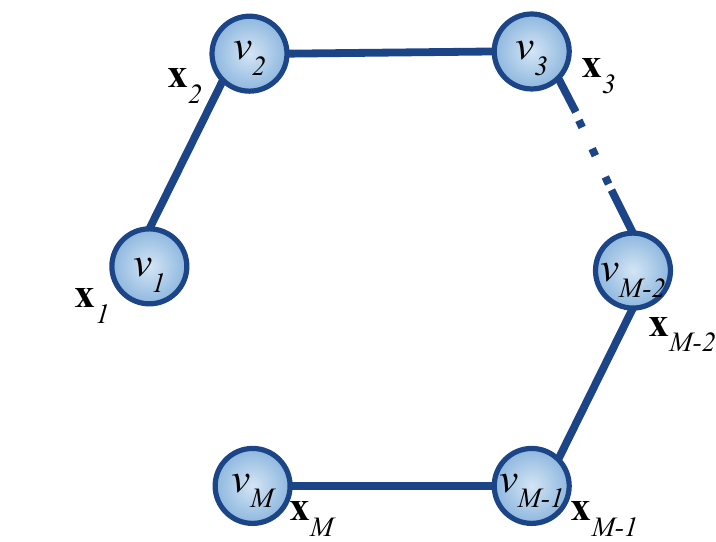}
    (c)
  \end{minipage} 
  \caption{Graph construction from speech input. (a) LLDs are extracted as node features $\mathbf{x}_i$ from raw speech segments; (b) cycle graph, and (c) chain graph.}
\label{fig:graphstr}
\end{figure}

Our graph construction strategy follows a simple frame-to-node transformation, where $M$ frames (short, overlapping segments) of the speech signal form the $M$ nodes in $\mathcal{G}$ (see Fig.~\ref{fig:graphstr}(a)). Since the graph structure is not naturally defined here, we investigate two simple undirected graph structures: (i) a \emph{cycle graph} defined by the adjacency matrix $\mathbf{A}_c$, and (ii) a \emph{line graph} defined by adjacency $\mathbf{A}_l$. The two graph structures are shown in Fig.~\ref{fig:graphstr}(b)-(c).
\begin{equation*}
    \mathbf{A}_c =\begin{bmatrix}
      0 & 1 & 0 & \cdots  & 1 \\
      1 & 0 & 1 & \cdots  & 0 \\
      0 & 1 & 0 & \cdots  & 0 \\
      \vdots & \vdots & \vdots & \ddots & \vdots \\
      1 & 0 & \cdots & 1 & 0  
     \end{bmatrix}
     \mathbf{A}_l =\begin{bmatrix}
      0 & 1 & 0 & \cdots  & 0 \\
      1 & 0 & 1 & \cdots  & 0 \\
      0 & 1 & 0 & \cdots  & 0 \\
      \vdots & \vdots & \vdots & \ddots & \vdots \\
      0 & 0 & \cdots & 1 & 0  
     \end{bmatrix}
\end{equation*}
These two graph structures are important because of the special structures of their graph Laplacians, which significantly simplifies spectral GCN operations. This is discussed in the following section in more detail.

Each node $v_i$ is also associated with a \textit{node feature} vector $\mathbf{x}_i \in \mathbb{R}^{P}$. A node feature vector contain LLDs extracted from the corresponding speech segment. A feature matrix ${\bf X}\in\mathbb{R}^{M\times P}$ containing all node feature vectors is defined as ${\bf X}=[\mathbf{x}_1,\cdots \mathbf{x}_M]$.
\subsection{Graph classification} \label{graph_class}
Given a set of (utterances transformed to) graphs $\{G_1, ..., G_N\}$ and their true labels $\{\mathbf{y}_1, ..., \mathbf{y}_N\}$ represented as one-hot vectors, our task is to develop a GCN architecture that is able to recognize the emotional content in the utterances. Our architecture comprises two graph convolution layers, a pooling layer that yields a graph-level embedding vector, followed by a fully connected layer that produces the discrete emotion labels (Fig.\ref{fig:graph_con}).

\vspace{1mm}
\noindent\textbf{Graph convolution layer.} We base our model on a spectral GCN, which performs graph convolution in the spectral domain. Following the theory of graph signal processing \cite{6494675}, graph convolution in time domain is defined as
\begin{equation}
    \mathbf{h} = \mathbf{x}_i\ast \mathbf{w}
\end{equation}
where $\bf w$ is the graph convolution kernel (learnable) and $\mathbf{x}_i$ is the input node features. 
This is equivalent to a product in the graph spectral domain.
\begin{equation}
    \hat{\mathbf{h}}= \hat{\mathbf{x}_i} \odot \hat{\mathbf{w}}
\end{equation}
 where $\hat{\mathbf{h}}$, $\hat{\mathbf{x}_i}$, and $\hat{\bf g}$ denote the output, node features and the convolution filter in the graph spectral domain i.e., their graph Fourier transforms (GFT). Considering the node feature matrix and sdopting a matrix notation, we get
\begin{equation}
    \hat{\mathbf{H}}= \hat{\mathbf{X}} \hat{\mathbf{W}}
\end{equation}
In order to have $\hat{\mathbf{X}}$ and $\hat{\bf W}$, we usually compute the normalized graph Laplacian matrix 
\begin{equation} \label{eq:norm_lap}
    \mathcal{L}= \mathbf{D}^{-\frac{1}{2}}\mathbf{L}\mathbf{D}^{-\frac{1}{2}}
\end{equation}
where $\mathbf{D}$ is the degree matrix, $\mathbf{L}=\mathbf{D} -\mathbf{A}$ where $\mathbf{A}$ is the adjacency matrix of the graph. 
The eigen decomposition of $\mathcal{L}$ can be written as
\begin{equation} 
    \label{eq:eigen_dec}
        \mathcal{L}= \mathbf{U}\Lambda\mathbf{U}^T = \displaystyle\sum_{i=1}^{M} \lambda_i\textbf{u}_i{\textbf{u}_i}^T
\end{equation}
where $\lambda_i$ is the $i^{th}$ eigen value of $\mathcal{L}$ corresponding to the eigen vector $\textbf{u}_i$, $\Lambda=\text{diag}(\lambda_i)$ and ${\bf U}= [\mathbf{u}_1, \mathbf{u}_2 \cdots \mathbf{u}_N]$. The exact graph convolution operation is thus defined as
\begin{align*} 
    \hat{\mathbf{H}}&=(\mathbf{U}^T\mathbf{X})(\mathbf{U}^T\mathbf{W})\\
     \mathbf{H}&=\mathbf{U}\hat{\mathbf{H}}
\end{align*}
\noindent The graph convolution propagation at $k^{th}$ layer thus becomes
\begin{equation} \label{eq:spec_conv_org}
\begin{aligned}
    \mathbf{H}^{(k+1)}=\mathbf{U}\Big((\mathbf{U}^T\mathbf{H}^{(k)})(\mathbf{U}^{T}\mathbf{W}^{(k)})\Big)
\end{aligned}
\end{equation}
where $\mathbf{H}^{(0)}= \mathbf{X}$ and $\bf W$ is learnable.
\noindent Note that for $\mathbf{A} = \mathbf{A}_c$ (cycle graph), $\bf L$ takes the following form
\begin{equation*}
    \mathbf{L} =\begin{bmatrix}
      2 & -1 & 0 & \cdots  & -1 \\
      -1 & 2 & -1 & \cdots  & 0 \\
      0 & -1 & 2 & \cdots  & 0 \\
      \vdots & \vdots & \vdots & \ddots & \vdots \\
      -1 & 0 & \cdots & -1 & 2  
     \end{bmatrix}
\end{equation*}
The $\bf L$ is circulant and GFT is equivalent to the Discrete Fourier Transform (DFT) \cite{ortega2018graph}. Similarly, for $\mathbf{A} = \mathbf{A}_l$ (line graph), GFT is equivalent to Discrete Cosine Transform (DCT). This makes the convolution operation convenient and computationally efficient as we can avoid eigen decomposition that can be computationally expensive for arbitrary graph structures.

Following a recent work on GCN \cite{xu2018how}, we propose to learn the convolution kernel in Eq.~\eqref{eq:spec_conv_org} in terms of a Multi-Layer Perceptron (MLP). Finally, our convolution operation takes the following form
\begin{equation} \label{eq:spec_conv}
\begin{aligned}
    \mathbf{H}^{(k+1)}=\mathbf{U}\Big(\mathrm{MLP}\big(\mathbf{U}^T\mathbf{H}^{(k)}\big)\Big), 
\end{aligned}
\end{equation}
where, only the MLP parameters are learnable.

\vspace{1mm}
\noindent\textbf{Pooling layer.} Our objective is to classify entire graphs (as opposed to the more popular task of graph node classification). Hence, we need a function to attain a \emph{graph-level} representation $\mathbf{h}_G\in\mathbb{R}^Q$ from the node-level embeddings. This can be obtained by pooling the node-level embeddings $\mathbf{H}^{(k)}$ at the final layer before passing them onto the classification layer. Common choices for pooling functions in graph domain are mean, max and sum pooling \cite{kipf2017semi, niepert2016learning}. Max and mean pooling often fail to preserve the underlying information about the graph structure while sum pooling has shown to be a better alternative \cite{xu2018how}. We use sum pooling to obtain the graph-level representation: 
\begin{equation} \label{eq:graph_emb}
    \mathbf{h}_G = \mathrm{sumpool}(\mathbf{H}^{(K)})=\displaystyle\sum_{i=1}^{M}{\mathbf{h}_i}^{(K)}
\end{equation}
\indent The pooling layer is followed by one fully-connected layer which produces the classification labels. Our GCN model is trained with the cross-entropy loss  $= -\displaystyle\sum_{n} \mathbf{y}_n \log \mathbf{\tilde{y}}_n$.
\section{Experiments}
\label{sec:Exp}
In this section, we present experimental results and analysis to evaluate the performance of the proposed GCN architecture.
\begin{table}[tb]
\caption{SER results and comparison on the IEMOCAP databases in terms of weighted accuracy (WA) and unweighted accuracy (UA).}
\label{tab:IEMO_result}
\vspace{-3mm}
 \renewcommand*{\arraystretch}{1.3}
\begin{center}
\begin{tabular}{l | c c}
\hline
\bf Model & \bf WA (\%) & \bf UA (\%) \\ 
\hline \hline
\multicolumn{3}{c}{\emph{Graph baselines}}\\
\hline
GCN \cite{kipf2017semi} &56.14 &52.36\\
PATCHY-SAN \cite{niepert2016learning} &60.34 & 56.27\\
PATCHY-Diff \cite{ying2018hierarchical} &63.23 & 58.71\\
\hline 
\multicolumn{3}{c}{\emph{SER models}}\\
\hline
Attn-BLSTM 2016 \cite{huang2016attention} &59.33 & 49.96\\
BLR 2017 \cite{ma2017speech} &62.54 &57.85 \\
RNN 2017 \cite{mirsamadi2017automatic} &63.50 & 58.80 \\
CRNN 2018 \cite{luo2018investigation} &63.98 &60.35 \\
SegCNN 2019 \cite{mao2019deep} &64.53 &\bf 62.34 \\
LSTM 2019 \cite{latif2019direct} &58.72 & - \\
CNN-LSTM 2019 \cite{latif2019direct} & 59.23 & - \\
\hline 
{\bf Ours} (cycle) & \bf 65.29 & \bf 62.27 \\
{\bf Ours} (line) &  64.69 & 61.14 \\
{\bf Ours} (cycle w/o MLP) &  64.19 & 60.31 \\
\hline
\end{tabular}
\end{center}
\vspace{-8mm}
\end{table}
\subsection{Database}
\vspace{-1mm}
We evaluate our model on two most popular SER databases:  IEMOCAP \cite{busso2008iemocap} and MSP-IMPROV \cite{busso2016msp}. For both databases, a single utterance may have multiple labels owing to different annotators. We consider only the label which has majority agreement. 

\indent The \textbf{IEMOCAP} database contains a total of 12 hours of data collected over 5 dyadic sessions with 10 subjects.  
To be consistent with previous studies, we used four emotion classes :\textit{anger}, \textit{joy}, \textit{neutral}, and \textit{sadness}. The final dataset contains a total of $4490$ utterances including $1103$ \emph{anger}, $595$ \emph{joy}, $1708$ \emph{neutral} and $1084$ \emph{sad}.

The \textbf{MSP-IMPROV} database contains utterances from 12 speakers collected across six sessions. The  dataset contains a total of $7798$ utterances including $792$ samples for \emph{anger}, $3477$ for \emph{neutral}, $885$ for \emph{sad} and $2644$ samples for \emph{joy}.
\subsection{Node features}
\vspace{-1mm}
We extract a set of low-level descriptors (LLDs) from the speech utterances as proposed for Interspeech2009 emotion challenge \cite{schuller2009interspeech} using the OpenSMILE toolkit \cite{eyben2013recent}. The feature set includes Mel-frequency cepstral coefficients (MFCCs), zero-crossing rate, voice probability, fundamental frequency (F0) and frame energy. For each sample, we use a sliding window of length $25$ms (with a stride length of $10$ms) to extract the LLDs locally. Each feature is then smoothed using a moving average filter, and the smoothed version is used to compute their respective first order delta coefficients. In addition, we also add spontaneity as a binary feature for IEMOCAP as this information is known to help SER \cite{mangalam2017learning}. The spontaneity information comes with the database. Altogether this yields node feature vectors of dimension $P=35$ for IEMOCAP and $P=34$ for MSP-IMPROV (no spontaneity feature).

\subsection{Implementation Details}
\vspace{-1mm}
Each speech sample produces a graph of $M=120$ nodes, where each node corresponds to a (overlapping) speech segment of length $25$ms. Padding is used to make the samples of equal length. The dimension of the graph embedding is set to $Q=64$. We perform a 5-fold cross-validation and report both average weighted  accuracy (WA) and unweighted accuracy (UA). All parameters and validation remain the same for both the databases. 
Our network weights are initialized following the Xavier initialization \cite{glorot2010understanding}. We used Adam optimizer with a learning rate of $0.01$ and a decay rate of $0.5$ after each $50$ epochs for all experiments. We used Pytorch for our experiments on an NVIDIA RTX-2080Ti GPU. 
%
%
\begin{table}[tb]
\caption{SER results and comparison on the MSP-IMPROV databases in terms of weighted accuracy (WA) and unweighted accuracy (UA).}
\label{tab:MSP_result}
\vspace{-3mm}
 \renewcommand*{\arraystretch}{1.3}
\begin{center}
\begin{tabular}{l | c c}
\hline
\bf Model & \bf WA (\%) & \bf UA (\%) \\ 
\hline \hline
\multicolumn{3}{c}{\emph{Graph baselines}}\\
\hline
GCN \cite{kipf2017semi} &54.71 &51.42\\
PATCHY-SAN \cite{niepert2016learning} &55.47 &52.33\\
PATCHY-Diff \cite{ying2018hierarchical} &56.18 &53.12\\
\hline 
\multicolumn{3}{c}{\emph{SER models}}\\
\hline
ProgNet 2017 \cite{GideonKADP17} &\bf 58.40 &-\\
CNN 2019 \cite{latif2019direct} &50.84 &-\\
LSTM 2019 \cite{latif2019direct} &51.21 &-\\
CNN-LSTM 2019 \cite{latif2019direct} & 52.36 &-\\
\hline 
{\bf Ours} (cycle) & \bf 57.82 & \bf 55.42\\
{\bf Ours} (line) &  57.08 &54.75\\
{\bf Ours} (cycle w/o MLP) &  56.82 &53.22\\
\hline
\end{tabular}
\end{center}
\vspace{-4mm}
\end{table}
\begin{table}[tb]
\caption{Model size comparison in terms of learnable parameters}
\label{tab:netsize}
 \renewcommand*{\arraystretch}{1.4}
 \vspace{2mm}
\begin{center}
\begin{tabular}{l c c l r}
\hline
GCN\ & PTCHY-SAN & PTCHY-Diff & BLSTM & \bf Ours \\
\hline
$\sim$76K & $\sim$60K & $\sim$68K & $\sim$0.8M & \bf $\sim$30K\\
\hline
\end{tabular}
\end{center}
\vspace{-4mm}
\end{table}

\subsection{Results and Analysis}
\textbf{Comparison with graph-based models.} We compare our model against three state-of-the-art deep graph models using the same node features and a cycle graph structure. 

\emph{GCN} \cite{kipf2017semi}. A natural baseline to compare with our model is a spectral GCN in its standard form. The original network \cite{kipf2017semi} is designed for node classification and only yields node-level embeddings. To obtain a graph-level embedding, we used the sum pooling function. 

\emph{PATCHY-SAN} \cite{niepert2016learning}. A recent architecture that learns CNNs for arbitrary graphs. This architecture is originally developed for graph classification.

\emph{PATCHY-Diff} \cite{ying2018hierarchical}. A recent work on hierarchical GCN proposes to use differentiable pooling layer between graph convolution layers. We used this pooling layer with PATCHY-SAN as in the original paper.

Table \ref{tab:IEMO_result} and Table \ref{tab:MSP_result} compare our model against these existing graph models (Graph baselines) in terms of SER accuracy. All these models use the same node features. Clearly, our model outperforms all the graph baselines by a significant margin. Compared to the popular GCN \cite{kipf2017semi}, our model improves the recognition accuracy by more than $9\%$ and $3\%$ on IEMOCAP and MSP-IMPROV respectively. This result indicates that accurate convolution in graph domain improves the accuracy significantly.

\vspace{1mm}\noindent\textbf{Comparison with SER state-of-the-art.} In addition to the graph baselines, we compare our model with a number of recent SER models. These include a Bayesian model \cite{ma2017speech}, CNN models (\cite{latif2019direct}, SegCNN \cite{mao2019deep}), Recurrent Neural Network (RNN) architectures (\cite{mirsamadi2017automatic}, RCNN \cite{luo2018investigation}), LSTM models (\cite{latif2019direct}, Attn-BLSTM \cite{huang2016attention}) and a CNN-LSTM model \cite{latif2019direct}. 
The majority of the above models (except the models by Latif et al. \cite{latif2019direct}) use LLDs as input. 

Tables \ref{tab:IEMO_result} and \ref{tab:MSP_result} show that our model outperforms (i) all graph baselines despite a simpler architecture and (ii) all LSTM-based architectures (a class of models most commonly used in SER) on both databases by a significant margin. Our model (with the cycle graph structure) achieves highest WA on IEMOCAP and comparable to the state-of-the-art WA on MSP-IMPROV. In terms of UA, our model's performance is comparable to SegCNN on IEMOCAP.
 Nevertheless, our model has significantly fewer parameters: $30$K learnable parameters vs.\ $8.8$M  in SegCNN and $0.8$M in LSTM.

\vspace{1mm}\noindent\textbf{Network size.} Table \ref{tab:netsize} compares the number of learnable network parameters for various models with ours. All graph networks are smaller (an order of magnitude smaller) than LSTM architectures yet highly accurate. Our model has the highest accuracy with half the parameters of other graph-based networks. This is owing to the light-weight convolution operation and because of the choice of our graph structure. In our approach graph structure remains the same for all samples, which requires us to compute the eigen-decomposition only once. This operation can even be replaced by directly using DFT or DCT kernels as mentioned earlier.
\begin{table}[tb]
\caption{Comparing different pooling strategies for our model on the IEMOCAP database.}
\label{tab:pool}
\vspace{2mm}
  \renewcommand*{\arraystretch}{1.2}
 \begin{center}
 \begin{tabular}{c | c c c c}
  \hline
{\bf Pooling}  & {Maxpool} & {Meanpool}  & {Sumpool} \\
 \hline
 {\bf WA (\%)}  &  {$61.68$}             &{$62.45$}  &  {$\mathbf{65.29}$}  \\
\hline
 \end{tabular}
\vspace{-4mm}
 \end{center}
 \end{table}

\vspace{1mm}
\noindent\textbf{Ablation:} Between the two graph structures we investigated, higher SER accuracy is achieved using the cyclic structure on both the databases. With the line graph, the model accuracy is slightly lower. We also compare our model's performance for different pooling strategies (used to compute graph-level representation from node embeddings) in Table \ref{tab:pool}. Aligned with observations made in past works (e.g., \cite{xu2018how}), sumpool shows improvement over meanpool and maxpool by $2.8\%$ and $3.6\%$ on IEMOCAP. When using graph convolution without MLP (see Eq. \ref{eq:spec_conv_org}) performance drops by $1\%$ (see Table \ref{tab:IEMO_result} and \ref{tab:MSP_result}). These results confirm that each proposed component in our network (MLP convolution, sumpool) contributes positively towards its performance.
\section{Conclusion}
\label{sec:conclusions}
We proposed a compact and efficient GCN architecture (with only 30K learnable parameters) for recognizing emotion content in speech. To the best of our knowledge, this is the first graph-based approach to SER. We transformed speech utterances to graphs with simple structures that largely simplify the convolution operation on graphs. Also, the graph structure we defined remains the same for all samples as our edges are not weighted. This leads to a light-weight GCN architecture which outperforms LSTMs, standard GCNs and several other recent graph models in SER. Our model produces higher or comparable performance to the state-of-the-art on two benchmark databases with significantly fewer learnable parameters. 

\balance
\bibliographystyle{IEEEbib}

\bibliography{mybib}
\end{document}